\documentclass[]{TEAI}
\usepackage{helvet}
\usepackage{xspace}
\usepackage{tikz}
\usepackage{pifont}

\definecolor{mydarkblue}{rgb}{0.21,0.49,0.74}
\def\eg{\textit{e.g.}}
\def\ie{\textit{i.e.}}

\newcommand*{\system}{RepWAM\@\xspace}

\newcommand{\keywords}[1]{\par\vspace{0.5em}\noindent{\small\textbf{Keywords:} #1}\par\vspace{0.5em}}
\newcommand{\projectlinks}[1]{\renewcommand{\contributionlist}{{\contributionfont #1}}}

\title{\system: World Action Modeling with Representation Visual-Action Tokenizers}

\author[1]{Junke Wang}
\author[2]{Qihang Zhang}
\author[2]{Shuai Yang}
\author[2]{Yiming Luo}
\author[2]{Yujun Shen}
\author[1\dagger]{Zuxuan Wu}
\author[1\dagger]{Yu-Gang Jiang}
\author[3,2\dagger]{Yinghao Xu}

\affiliation[1]{Institute of Trustworthy Embodied AI, Fudan University}
\affiliation[2]{Robbyant, Ant Group}
\affiliation[3]{Hongkong University of Science and Technology}
\projectlinks{Web page: \href{https://wdrink.github.io/RepWAM}{https://wdrink.github.io/RepWAM}\\ Code: \href{https://github.com/wdrink/RepWAM}{https://github.com/wdrink/RepWAM}}

\abstract{\leavevmode This work presents \system, a world action model (WAM) built with representation visual-action tokenizers. Existing WAMs typically inherit reconstruction-oriented video tokenizers from pretrained video generation models. While effective at preserving visual fidelity, pixel reconstruction alone provides limited guidance for learning the instruction-following dynamics that bridge future prediction and robot control. To address this, we explore the semantic visual-action latent space for representation-centric world action modeling. Specifically, we first train a representation visual-action tokenizer that compresses the visual inputs as aligned visual and latent action tokens. With this, we pretrain our WAM to jointly model future visual states and the latent actions that connect them under language instructions, and subsequently adapt it on real robot trajectories for closed-loop manipulation. Experiments on real-world manipulation tasks and simulation benchmarks show that \system delivers strong performance compared with existing vision-language-action models and WAMs. These results establish representation visual-action tokenization as a promising foundation for world action models and a step toward generalist robot policies.

}

\begin{document}
\maketitle
\begingroup
\renewcommand{\thefootnote}{\fnsymbol{footnote}}
\footnotetext[1]{Correspondence to Zuxuan Wu, Yu-Gang Jiang, and Yinghao Xu.}
\endgroup

\section{Introduction}
\label{sec:intro}
The capability of an agent to model dynamics and derive behaviors is fundamentally determined by the representation through which it perceives the world~\cite{dosovitskiy2020image,bolya2026perception}. This matters most for world action models (WAMs)~\cite{wang2026world,ye2026world,li2026causal}, which extend classical world modeling~\cite{ha2018world,ha2018recurrent,ali2025world,agarwal2025cosmos} from passive future prediction to embodied control: a single latent space must both forecast visual dynamics and infer the robot actions that realize them. In current WAMs, however, this representation is often borrowed from pretrained video generators~\cite{wan2025wan} that are optimized for pixel reconstruction rather than visual semantics or robot action.

This work asks: what kind of representation does a world action model actually need? Current WAM representations fall short along two distinct dimensions. On the visual side, reconstruction-driven video tokenizers~\cite{wang2024omnitokenizer,wan2025wan} spend latent capacity on low-level appearance such as background texture, while object identity, spatial relations, and interaction cues that drive manipulation are underrepresented. This leaves instruction-conditioned world modeling poorly grounded: language goals about specific objects or interactions must be inferred from a latent space dominated by appearance variation rather than task-relevant semantics. On the action side, visual latents and motor commands reside in disjoint spaces, forcing an inverse dynamics model (IDM) to bridge this modality gap at every step~\cite{ye2025latent}. Together, these failures leave WAM representations visually shallow and structurally decoupled from action. 

Motivated by this, this work revisits the design of the latent space underlying world action models and proposes \system, a representation-centric WAM built upon representation visual-action tokenizers. At its core, \system introduces a semantic video tokenizer by aligning the latent space of a video autoencoder with a frozen visual foundation model~\cite{bolya2026perception}. Building on these semantically rich visual latents, we further learn a latent action tokenizer that captures manipulation-centric motions tightly coupled to object-level interactions. Together, the two tokenizers form a unified semantic visual-action tokenizer that narrows the modality gap between visual latents and robot actions. On top of this space, we pretrain a world action model that jointly forecasts future visual states and their corresponding latent actions, and then adapt it to real robot data for downstream manipulation.

We evaluate \system on both real-world manipulation tasks and simulation benchmarks. Compared with vision-language-action (VLA) and WAN-pretrained WAM baselines, \system delivers competitive closed-loop behavior and reaches \textbf{89.3} on Easy tasks and \textbf{88.4} on Hard tasks in RoboTwin 2.0. These results suggest that grounding visual and action latents in a shared semantic representation is a promising foundation for world action modeling.
\section{Related Work}
\label{sec:related}

\subsection{World Action Models} 
World action models extend visual world modeling to robot control by using predicted scene evolution as an intermediate representation for action generation. Motus~\cite{bi2025motus} introduces a mixture-of-transformer architecture that assigns understanding, video generation, inverse dynamics, and action prediction to specialized experts within a unified latent action world model. Lingbot-VA~\cite{li2026causal} formulates WAM training through causal world modeling, jointly learning frame prediction and policy execution with an autoregressive diffusion model that captures how motor decisions lead to future visual changes. Along a related direction, DreamZero~\cite{ye2026world} initializes from a pretrained video generation model and adapts it for joint video-action modeling, demonstrating strong zero-shot generalization to unseen tasks, scenes, and embodiments. These systems highlight the value of predictive visual dynamics, but they also leave open whether explicit future generation is necessary once the policy is deployed. Fast-WAM~\cite{yuan2026fast} disentangles training-time video modeling from inference-time future generation: it skips future prediction at test time yet remains competitive with imagine-then-execute variants, indicating that video supervision can improve policy learning even without online imagination. In this work, we explore a complementary direction: constructing semantic visual tokens and aligned latent action tokens as the representation space on which world action modeling is trained.

\subsection{Latent Action Models} 
Latent action models recover action-like variables from observation sequences, making unlabeled videos usable for downstream control. Genie~\cite{bruce2024genie} establishes this direction in generative environments, where latent actions make video dynamics controllable without ground-truth action labels. LAPO~\cite{schmidt2024learning} moves from controllable generation toward policy learning, recovering latent-action policies and dynamics models from observed transitions before adapting them with limited supervision or reward. LAPA~\cite{ye2025latent} scales this principle to robot manipulation by learning discrete latent actions between image frames and using them as pretraining targets for VLA models. Moto~\cite{chen2025moto} shifts the abstraction from generic action codes to motion tokens, using autoregressive video pretraining to transfer motion regularities into robot policies. This line of work progressively turns visual change into a substitute for missing motor labels. In this work, we instead learn latent actions within a semantic visual latent space, making each action token a transformation between semantic visual states rather than an isolated code extracted from raw pixel changes. This design aligns the action abstraction with the visual tokens used by the world model, making it better suited for joint visual-action modeling in WAMs.

\section{Method}
\label{sec:method}

\begin{figure}[!t]
\centering
\includegraphics[width=\linewidth]{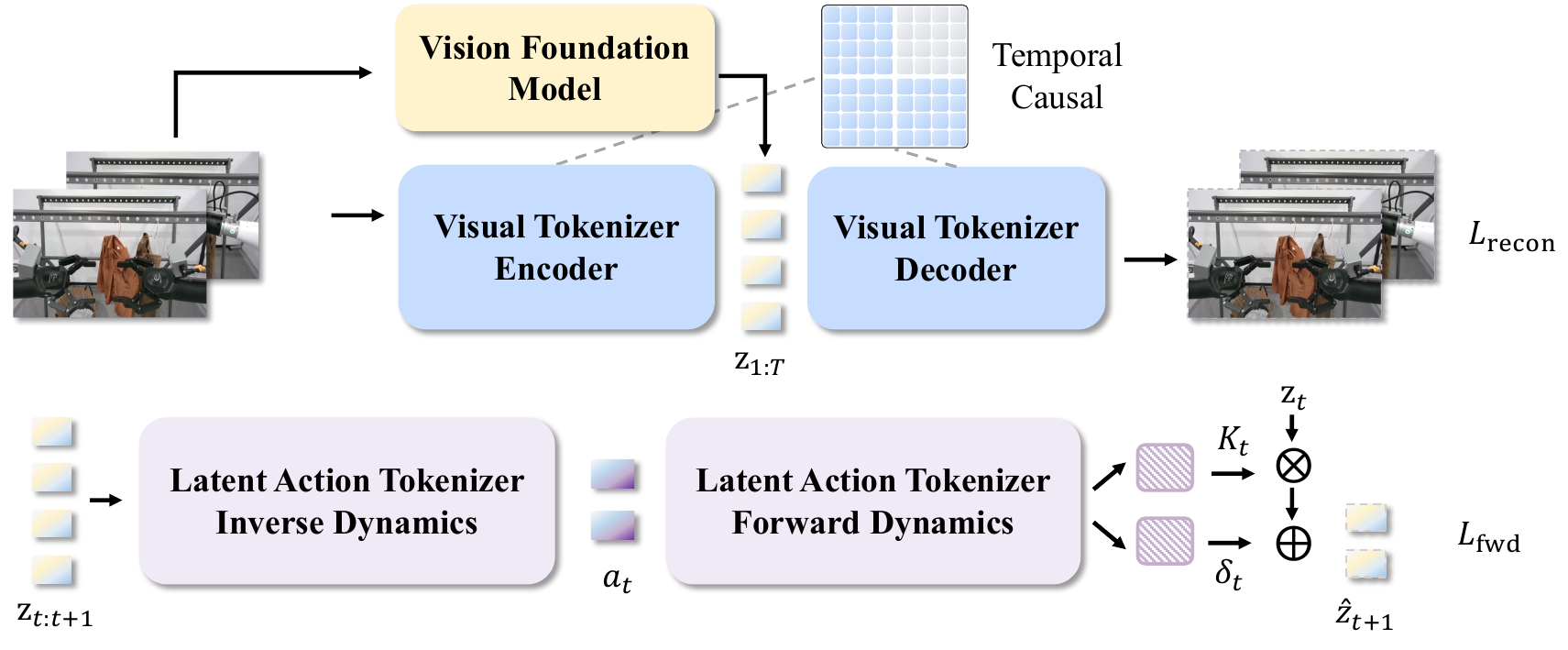}
\caption{Overview of our representation visual-action tokenizer, which aligns visual latents with a frozen visual foundation model to obtain semantically rich visual tokens, and induces latent actions as transitions within this shared semantic space via coupled inverse and forward dynamics models.}
\label{fig:network}
\end{figure}

\subsection{Overview}
\label{subsec:overview}

Given visual observations $o_{1:T}$, a language instruction $y$, and robot actions $a_{1:T-1}$, world action models (WAMs)~\cite{li2026causal,ye2026world} factor manipulation into a two-stage causal process: the world model expert $p_{\theta_1}$ forecasts the next observation $\hat{o}_{t}\sim p_{\theta_1}(o_t\mid o_{<t},y)$, and the action expert $p_{\theta_2}$ infers the action $\hat{a}_{t-1}\sim p_{\theta_2}(a_{t-1}\mid o_{<t},\hat{o}_{t},y)$ that produces this transition, typically realized by an inverse dynamics model (IDM)~\cite{du2023learning}. In practice both stages operate over latents $z_{1:T}$ produced by a video tokenizer~\cite{wan2025wan} rather than raw pixels.

In this work, we posit that the fidelity of this formulation is bottlenecked by the latent spaces in which $z_t$ and $a_{t-1}$ reside. Specifically, the visual latent $z_t$ is inherited from reconstruction-oriented tokenizers tuned to appearance rather than semantics, making it challenging for the world model expert to learn instruction-following dynamics. Meanwhile, $a_{t-1}$ lives in a space decoupled from $z_t$, forcing the action expert to bridge a modality gap at every step.

Our goal is to overcome these limitations by establishing a semantic latent space in which visual states and their action-induced transitions are jointly represented. To this end, we present \system, a representation-centric WAM built around a representation visual-action tokenizer (RepViTok) that aligns visual latents with a visual foundation model and induces latent action tokens as transitions within the same space (Sec.~\ref{subsec:tokenizer}). Built on these aligned tokens, \system pretrains a causal world action model via flow matching over paired world model and action experts, and adapts it to robot demonstrations for executable control (Sec.~\ref{subsec:causal_wam}). The overall framework is illustrated in Figure~\ref{fig:network}.

\subsection{Representation Visual-Action Tokenizer}
\label{subsec:tokenizer}

To close the latent-space gap above, we design a tokenizer that yields semantically aligned visual tokens together with action tokens that capture transitions between them. Taking the visual observations $o_{1:T}$ as input, our representation visual-action tokenizer produces visual tokens $z\in\mathbb{R}^{T'L\times d_v}$ that compactly encode visual content and latent action tokens $\ell\in\mathbb{R}^{(T'-1)\times d_\ell}$ that capture temporal dynamics, where the visual tokens are organized as $T' = 1 + \frac{(T-1)}{4}$ latent frames of $L$ spatial tokens each after temporal patchification. The two tokenization steps are performed sequentially, so latent actions are induced from the learned visual latent space rather than directly from pixels.

\paragraph{Visual tokenization.}
The visual tokenizer is a vision transformer (ViT)~\cite{dosovitskiy2020image,wang2024omnitokenizer} autoencoder. The initial frame $o_1$ and the subsequent frames $o_{2:T}$ are split into $16\times 16$ patches and $4\times 16\times 16$ (temporal$\times$height$\times$width) tubelets, respectively. We concatenate the resulting tokens along the sequence dimension and feed them into the encoder $E_\theta$, which consists of stacked attention blocks with temporal-causal masking across frames and full spatial attention within each frame. The encoder output is then passed through an attention-based projection layer followed by layer normalization to obtain the video latents $z$. The decoder $D_\theta$ follows a symmetric architecture and maps the latents back to pixels through an unpatchify head composed of transposed convolution layers.

We supervise the visual tokenizer with a reconstruction objective that combines multiple losses:
\begin{equation}
\mathcal{L}_{\mathrm{rec}} = \lambda_1 \lVert o-\hat{o}\rVert_1 + \lambda_{\mathrm{perc}}\mathcal{L}_{\mathrm{perc}}(o,\hat{o}) + \lambda_{\mathrm{gan}}\mathcal{L}_{\mathrm{gan}}(\hat{o}),
\end{equation}
where $\hat{o}{=}D_\theta(z)$ denotes the reconstructed frames, $\mathcal{L}_{\mathrm{perc}}$ and $\mathcal{L}_{\mathrm{gan}}$ are the perceptual and adversarial losses, and $\lambda_1$, $\lambda_{\mathrm{perc}}$, $\lambda_{\mathrm{gan}}$ weight the three terms.

Beyond reconstruction, we add a feature-alignment loss that pulls the video latents toward a frozen visual foundation model~\cite{bolya2026perception}. Let $G$ denote this teacher model and let $W_{\mathrm{align}}$ be a linear projection layer that matches the teacher dimension. The alignment objective is
\begin{equation}
\mathcal{L}_{\mathrm{align}} = \left\lVert W_{\mathrm{align}} z - \mathrm{avg}(G(o))\right\rVert_2^2 .
\end{equation}

where $\mathrm{avg}$ denotes temporal average pooling following Perception Encoder~\cite{bolya2026perception}.

The visual tokenizer is jointly supervised by the total objective $\mathcal{L}_{\mathrm{vis}}=\mathcal{L}_{\mathrm{rec}}+\lambda_{\mathrm{align}}\mathcal{L}_{\mathrm{align}}$, where $\lambda_{\mathrm{align}}$ balances reconstruction fidelity and semantic alignment.

\paragraph{Latent action tokenization.}
Building on these semantically aligned visual tokens, we then train action tokens in the same latent space, so that each action token represents a transition between two visual states. We freeze the visual tokenizer and train a latent action tokenizer (LAT) that couples an IDM $q_\phi$ with a forward dynamics model (FDM) $f_\psi$. 

For consecutive latent frames $z_t,z_{t+1}\!\in\!\mathbb{R}^{L\times d_v}$, the IDM compresses the transition into $\ell_t\!\in\!\mathbb{R}^{d_\ell}$ ($d_\ell\ll d_v$), preventing content leakage. The FDM then realizes $\ell_t$ as a transport map $K_t\!\in\!\mathbb{R}^{L\times L}$ and a residual $\delta_t\!\in\!\mathbb{R}^{L\times d_v}$:
\begin{equation}
\begin{aligned}
\ell_t &= q_\phi(z_t,z_{t+1}), \\
K_t, \delta_t &= f_\psi(z_t,\ell_t), \\
\hat{z}_{t+1} &= K_t\, z_t + \delta_t,
\end{aligned}
\end{equation}
where $K_t\, z_t$ denotes left-multiplication along the spatial-token dimension of $z_t$, and $\hat{z}_{t+1}$ is the reconstruction of the next visual latent. Intuitively, $K_t$ acts as a soft transport operator inspired by optical flow~\cite{teed2020raft} in the semantic token space, routing visual content according to the state change induced by $\ell_t$, while $\delta_t$ captures residual changes that cannot be explained by transport alone. Since this transformation is defined over visual states rather than embodiment-specific motor coordinates, the resulting latent action describes a transferable task-level transition.

The LAT is trained with a forward next-latent prediction loss and a backward consistency loss, where the backward $\hat{z}_t$ is obtained by running the LAT on the reversed pair $(z_{t+1},z_t)$:
\begin{equation}
\mathcal{L}_{\mathrm{fwd}} = \sum_{t=1}^{T'-1} \bigl\lVert \hat{z}_{t+1}-z_{t+1}\bigr\rVert_2^2,
\qquad
\mathcal{L}_{\mathrm{cons}} = \sum_{t=1}^{T'-1} \bigl\lVert \hat{z}_{t}-z_t\bigr\rVert_2^2 .
\end{equation}

\subsection{Causal World Action Models}
\label{subsec:causal_wam}

With aligned visual and action tokens in hand, we now train the \system world action model that generates both streams jointly under language conditioning.

\paragraph{Causal diffusion transformer.}
We cast world action modeling as causal generation over visual-action chunks. The language instruction $y$ is embedded by a pretrained text encoder into conditioning tokens $c$. Given a chunk size $k$, each chunk groups the visual latents and latent actions over a short temporal window:
\begin{equation}
\begin{aligned}
u_{t:t+k} = \bigl[z_{t:t+k},\, \ell_{t:t+k-1}\bigr],
\\
s = \bigl[c,\, z_1,\, u_{t_1:t_1+k},\, \ldots,\, u_{t_N:t_N+k}\bigr],
\end{aligned}
\end{equation}
where $z_{t:t+k}$ are the flattened visual tokens in window $t\!:\!t{+}k$, $\ell_{t:t+k-1}$ the paired latent actions between consecutive visual steps, $t_1,\ldots,t_N$ the start indices of the $N$ chunks, and $s$ prefixes the chunk sequence with language tokens and the initial visual context $z_1$. For a target chunk $u_{t:t+k}$, $s_{<t}$ denotes this prefix together with all preceding chunks. A block-causal mask lets each chunk attend to $s_{<t}$ but not to future chunks~\cite{wang2026omnigen}. Within each block, visual and action tokens share attention weights but use modality-specific feed-forward networks (FFNs).

\paragraph{Flow-matching objective.}
We train the transformer with teacher forcing under a conditional flow-matching objective applied jointly to the world model and action experts. At each step we sample a Gaussian noise tensor $\epsilon_{t:t+k}\sim\mathcal{N}(0,I)$ with the same shape as the chunk and a time scalar $\alpha\sim\mathcal{U}(0,1)$, and form the linear interpolant:
\begin{equation}
x_\alpha = (1-\alpha)\epsilon_{t:t+k} + \alpha u_{t:t+k},
\qquad
\dot{x}_\alpha = u_{t:t+k}-\epsilon_{t:t+k},
\end{equation}
where $\dot{x}_\alpha$ is the target velocity that the network should regress. Using the velocity estimate $F_\theta(x_\alpha,\alpha,s_{<t})$ from the network, the training loss is:
\begin{equation}
\mathcal{L}_{\mathrm{FM}}
=
\mathbb{E}
\left[
\left\lVert F_\theta^{v}(x_\alpha,\alpha,s_{<t})-\dot{x}_{\alpha}^{v}\right\rVert_2^2
+
\lambda_a
\left\lVert F_\theta^{a}(x_\alpha,\alpha,s_{<t})-\dot{x}_{\alpha}^{a}\right\rVert_2^2
\right],
\end{equation}
where the superscripts $v$ and $a$ index the visual and latent-action components of the chunk, respectively, and $\lambda_a$ balances their contributions.

\section{Experiments}
\label{sec:result}

\subsection{Implementation Details}
\label{subsec:impl}

\noindent \textbf{Training data.} We train our representation visual-action tokenizer, RepViTok, on Panda-70M~\cite{chen2024panda}. We pretrain the WAM on AgiBot~\cite{bu2025agibot} with roughly $100$\,G video-action latent tokens whose action components are produced by our latent action tokenizer. For embodiment adaptation, we use a mixed real-robot corpus spanning AgiBot, RoboMIND~\cite{wu2024robomind}, RoboCOIN~\cite{wu2025robocoin}, and InternA1~\cite{tian2025interndata} with roughly $300$\,G tokens, where each demonstration provides continuous motor commands (\eg, end-effector motion and gripper control) aligned with the visual trajectory.

\noindent \textbf{Model and optimization.} Our visual autoencoder adopts $12$ transformer layers with hidden dimension $768$ for both the encoder and decoder, with reconstruction weights $\lambda_1{=}1$, $\lambda_{\mathrm{perc}}{=}1$, $\lambda_{\mathrm{gan}}{=}0.1$, and alignment weight $\lambda_{\mathrm{align}}{=}1$, following previous work~\cite{wang2024omnitokenizer}. The latent action IDM $q_\phi$ and FDM $f_\psi$ are both four-layer MLPs with hidden size $256$. Visual-latent and latent-action token dimensions are set to $d_v{=}96$ and $d_\ell{=}4$, respectively.

We train the causal WAM backbones at two scales, $1.3$B and $5$B, entirely from scratch. Both follow the same architecture: the world model expert is a causal diffusion transformer with $30$ layers and hidden dimension $h_v$ ($h_v{=}1536$ for the $1.3$B model and $h_v{=}3072$ for the $5$B model), while the action expert shares the same depth with a reduced hidden dimension $h_a{=}768$, contributing roughly $350$M additional parameters. We share the attention weights for both experts, but adopt independent feed-forward networks for each modality. Language instructions are encoded by a frozen PLM text encoder~\cite{cho2026perceptionlm} and injected via cross-attention. We optimize the WAM with the Muon optimizer~\cite{jordan2024muon}, peak learning rate $1{\times}10^{-2}$, weight decay $0.01$, and cosine annealing with linear warmup, in bfloat16 mixed precision with gradient clipping at $2.0$. The action loss weight is set to $\lambda_a{=}1$, and a uniform SNR sampler is applied to both experts. Following~\cite{li2026causal}, we sample the chunk size from $[1,4]$ during training, and set it to 2 during inference. The $1.3$B model packs episodes into sequences of up to $200$K tokens and is trained on $64$ H20 GPUs, whereas the $5$B model packs up to $160$K tokens on $128$ H20 GPUs.

\subsection{Real-World Experiments}
\label{subsec:real}

We evaluate \system on a Franka dual-arm robot platform across three manipulation tasks: (1) \textbf{\textit{Pick the fruits and put them into the plate}} requires localizing target objects among clutter and executing reliable top-down grasps under varied poses and object instances, followed by precise placement within a designated container. (2)
\textbf{\textit{Push the drawer and put the building block into it}} poses a long-horizon challenge: the policy must apply the correct handle motion to actuate the drawer, retrieve a small block from the scene, and deposit it inside without collision. (3) \textbf{\textit{Insert the test tube into the test tube rack}} demands fine-grained spatial alignment, requiring accurate tip localization relative to the receptacle and compliant contact management under tight geometric clearance. For each task, we fine-tune on 50 real-world demos for 500 steps with learning rate 1e-5 and sequence length 150K.

\begin{figure}[!t]
\centering
\includegraphics[width=0.8\linewidth]{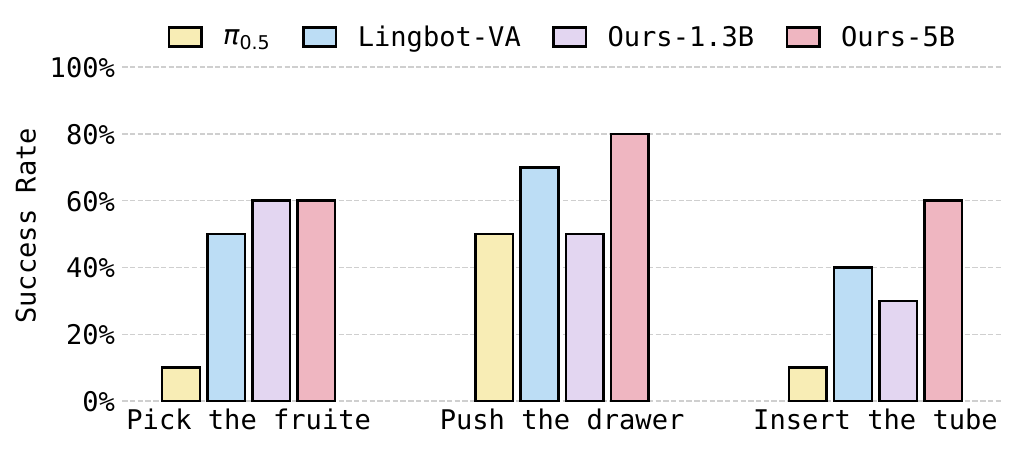}
\vspace{-0.1in}
\caption{Real-world success rate ($10$ rollouts per task) on three manipulation tasks, comparing $\pi_{0.5}$~\cite{black2025pi_} and Lingbot-VA~\cite{li2026causal} against our $1.3$B and $5$B WAMs.}
\label{fig:real_robots}
\end{figure}

We evaluate each task over $10$ physical rollouts and report success rate in Figure~\ref{fig:real_robots} for two model sizes, \system-$1.3$B and \system-$5$B. On the short-horizon \emph{pick-the-fruit} task, both sizes reach $60\%$, improving over $\pi_{0.5}$ ($10\%$) by $50$ points and over Lingbot-VA ($50\%$) by $10$ points, with the two tied since the bottleneck here is perception and grasping rather than model capacity. On the long-horizon \emph{push-the-drawer} task, \system-$5$B reaches $80\%$ (best overall, $+30$ over $\pi_{0.5}$, $+10$ over Lingbot-VA) and \system-$1.3$B reaches $50\%$, showing that multi-step articulated control benefits substantially from added capacity. The fine-grained \emph{insert-the-tube} task is the most demanding, where \system-$5$B reaches $60\%$ ($+50$ over $\pi_{0.5}$, $+20$ over Lingbot-VA) and \system-$1.3$B reaches $30\%$, again confirming that performance improves with model size. Across tasks, \system-$5$B is best or tied-best on every task, with the largest gains on long-horizon and fine-grained settings where the world model expert must roll out coherent visual dynamics over many steps before the action expert acts.

\begin{figure*}[!ht]
\centering
\includegraphics[width=\textwidth]{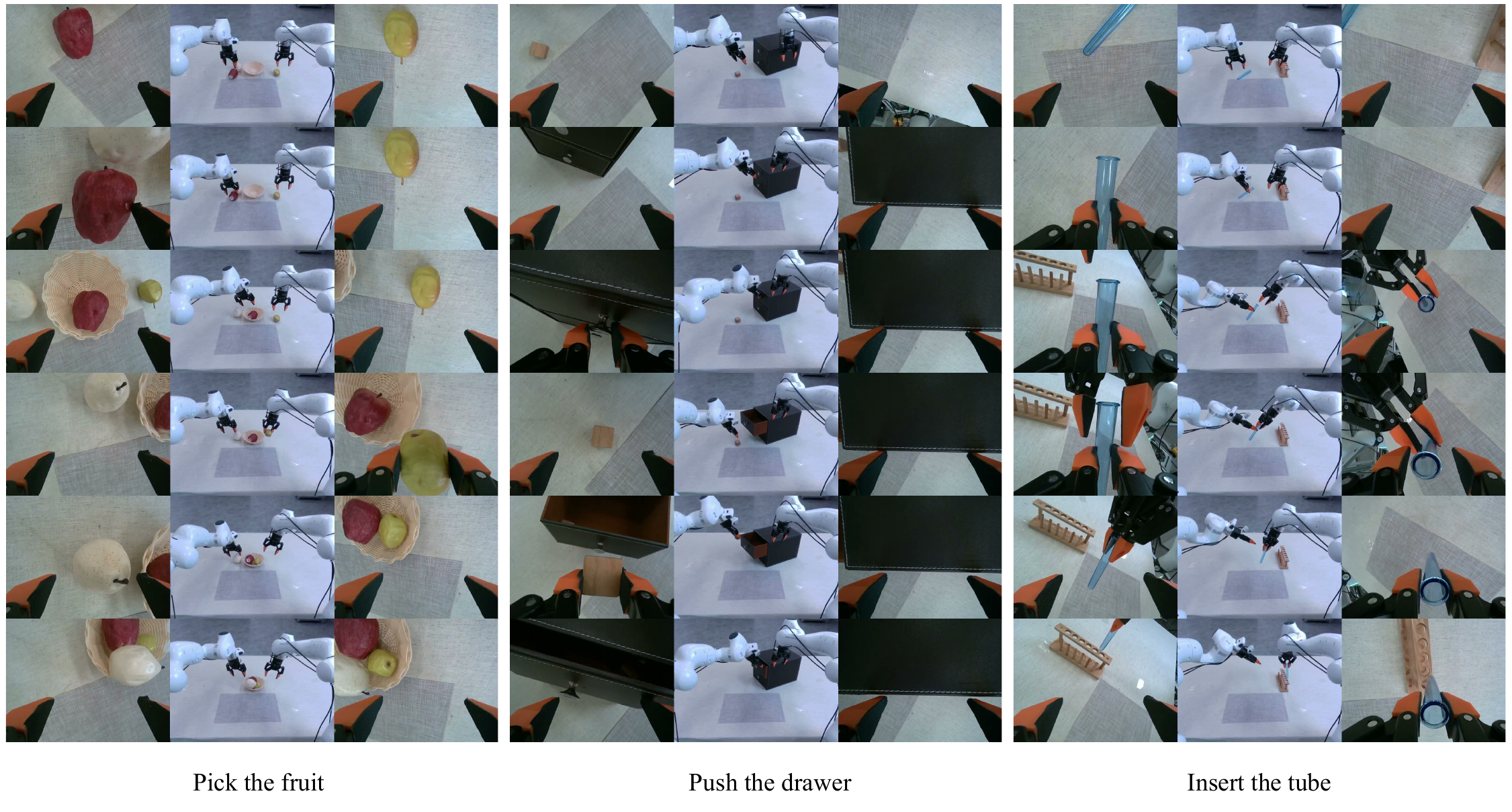}
\caption{Successful real-robot executions. From left to right, we show representative rollouts for \textit{picking the fruit}, \textit{pushing the drawer} and \textit{inserting the tube}.}
\label{fig:real_robot_vis}
\end{figure*}

Figure~\ref{fig:real_robot_vis} provides representative successful real-robot executions. From left to right, the visualizations cover \textit{picking the fruit}, \textit{pushing the drawer}, and \textit{inserting the tube}. In successful rollouts, the robot approaches target objects with stable grasp poses, follows the intended transport or articulation motion, and places or inserts objects near the desired goal without large corrective oscillations. We attribute this behavior to the representation visual-action tokenizer: RepViTok preserves object identity, contact-relevant geometry, and task context in the visual tokens, while the latent actions describe action-induced transitions in the same semantic space. As a result, the world action model can predict coherent scene evolution and the accompanying action stream before the adapted action expert maps these transitions to executable motor commands.

\subsection{Simulation Experiments}

\noindent \textbf{RoboTwin 2.0}~\cite{chen2025robotwin} is a large-scale dual-arm simulation benchmark with extensive domain randomization over scene composition, lighting conditions, camera viewpoints, and object physics. We evaluate on its standard task suite and report success rate under the official randomization settings.

As shown in Table~\ref{tab:robotwin}, \system-$5$B outperforms $\pi_{0.5}$ and Motus on the 50-task average across both Easy and Hard settings, demonstrating that semantic visual-action tokenization provides strong generalization for bimanual manipulation under randomized scenes. We hypothesize that the remaining performance gap to Lingbot-VA mainly comes from its use of WAN video-generation pretraining, whereas our WAM is trained from scratch without using pretrained weights. In the following ablation study, we further show that RepViTok improves over the WAN2.2 VAE used by such WAN-pretrained pipelines. These results suggest that the semantic structure introduced by representation-centric visual-action tokenization is a key factor for world action modeling, and can provide strong performance even without inheriting a pretrained video-generation backbone.

\begin{table*}[!ht]
\centering
\vspace{-0.05in}
\caption{Comparison on RoboTwin 2.0 benchmark. We follow Lingbot-VA~\cite{li2026causal} to categorize 50 tasks into different horizons.}
\label{tab:robotwin}
\vspace{0.05in}
\small
\renewcommand{\arraystretch}{1.5}
\setlength{\tabcolsep}{3pt}
\begin{tabular*}{\textwidth}{@{\extracolsep{\fill}}lcccccccccc@{}}
\hline
\multirow{2}{*}{\textbf{Metric}}
& \multicolumn{2}{c}{$\pi_{0.5}$~\cite{black2025pi_}}
& \multicolumn{2}{c}{Motus~\cite{bi2025motus}}
& \multicolumn{2}{c}{Lingbot-VA~\cite{li2026causal}}
& \multicolumn{2}{c}{Ours 1.3B}
& \multicolumn{2}{c}{Ours 5B} \\
\cline{2-3}\cline{4-5}\cline{6-7}\cline{8-9}\cline{10-11}
& Easy & Hard
& Easy & Hard
& Easy & Hard
& Easy & Hard
& Easy & Hard \\
\hline
Backbone pretrained
& \multicolumn{2}{c}{\checkmark}
& \multicolumn{2}{c}{\checkmark}
& \multicolumn{2}{c}{\checkmark}
& \multicolumn{2}{c}{\ding{55}}
& \multicolumn{2}{c}{\ding{55}} \\
\hline
Average$_{\mathrm{hor}=2}$
& 79.3 & 73.0
& 85.2 & 80.9
& 85.3 & 86.9
& 85.7 & 84.0
& 87.4 & 87.6 \\
Average$_{\mathrm{hor}=3}$
& 78.6 & 67.4
& 85.0 & 84.2
& 89.6 & 90.6
& 92.0 & 85.4
& 88.0 & 90.4 \\
\hline
Average$_{\mathrm{50\ Tasks}}$
& 82.7 & 76.8
& 88.7 & 87.0
& 92.9 & 91.6
& 86.6 & 83.1
& 89.3 & 88.4 \\
\hline
\end{tabular*}
\vspace{-0.2in}
\end{table*}

\subsection{Ablation Studies}

We adopt the $1.3$B WAM for ablation experiments, and train the model for $40{,}000$ steps on $32$ H20 GPUs. We split AgiBot into a training set, a seen evaluation set covering tasks observed during training, and an unseen evaluation set covering held-out tasks. We report three groups of metrics: (1) visual generation quality measured by gFVD, PSNR, and SSIM, (2) open-loop action accuracy measured by the open loop score (OLS)~\cite{li2026robointer} under threshold $0.03$, and (3) closed-loop execution on the real-robot pick-the-fruit task.

\noindent \textbf{Semantic visual tokenizer for world modeling.} We first study the effect of visual latent space on downstream world action modeling in Table~\ref{tab:tokenizer_ablation}. Our RepViTok achieves the strongest overall performance, reducing gFVD by $9.5$\% / $13.2$\% compared to the reconstruction-only WAN2.2 VAE~\cite{wan2025wan} and improving OLS from $13.68$ / $11.21$ to $18.82$ / $14.15$. The real-robot evaluation shows the same trend in closed-loop control: RepViTok reaches $30\%$ success on PickFruit, compared with $20\%$ for WAN2.2 VAE and $10\%$ for ViTok. ViTok improves OLS over WAN2.2 VAE but does not translate into reliable execution, suggesting that open-loop action accuracy alone is insufficient for robust manipulation. These results support our hypothesis that a latent space optimized solely for appearance deprives the world model expert of the semantic structure required to follow instructions, and prevents the action expert from grounding motor commands in scene content.

\begin{table*}[!ht]
\centering
\caption{Ablation study on tokenizer designs. We compare reconstruction-oriented and semantics-aware video tokenizers on AgiBot Eval Seen and Eval Unseen.}
\label{tab:tokenizer_ablation}
\vspace{0.02in}
\small
\renewcommand{\arraystretch}{1.2}
\begin{tabular*}{\textwidth}{@{\extracolsep{\fill}}lcccc|cccc|c@{}}
\hline
\multirow{2}{*}{\textbf{Model}}
& \multicolumn{4}{c|}{Eval Seen} & \multicolumn{4}{c|}{Eval Unseen} & \multirow{2}{*}{PickFruit} \\
\cline{2-5}\cline{6-9}
& gFVD & PSNR & SSIM & OLS 
& gFVD & PSNR & SSIM & OLS & \\
\hline
WAN2.2 VAE~\cite{wan2025wan}
& 67.42 & 17.34 & 0.67 & 13.68
& 83.98 & 16.86 & 0.64 & 11.21 & 20\% \\
ViTok
& 69.23 & 17.21 & 0.68 & 16.29
& 80.14 & 17.19 & 0.67 & 13.81 & 10\% \\
\textbf{RepViTok}
& \textbf{61.01} & \textbf{18.47} & \textbf{0.70} & \textbf{18.82}
& \textbf{72.91} & \textbf{17.72} & \textbf{0.67} & \textbf{14.15} & \textbf{30\%} \\
\hline
\end{tabular*}
\end{table*}

As discussed above, we hypothesize that the remaining gap to Lingbot-VA~\cite{li2026causal} mainly comes from its WAN video-generation pretraining. To examine this, we isolate the visual tokenizer while keeping the $1.3$B WAM setting fixed. As shown in Table~\ref{tab:robotwin_vae_comparison}, replacing WAN2.2 VAE with RepViTok improves the average success rate from $78.0$ to $86.6$ on Easy and from $76.0$ to $83.1$ on Hard, supporting the importance of semantics-aware visual tokenization for world action modeling.

\begin{table*}[!ht]
\centering
\caption{Comparison between different VAEs for the $1.3$B WAM on RoboTwin 2.0 simulation, Easy and Hard, 50 tasks.}
\label{tab:robotwin_vae_comparison}
\vspace{0.02in}
\small
\renewcommand{\arraystretch}{1.2}
\setlength{\tabcolsep}{3pt}
\begin{tabular*}{\textwidth}{@{\extracolsep{\fill}}lcccc@{}}
\hline
\multirow{2}{*}{\textbf{Metric}}
& \multicolumn{2}{c}{WAN2.2 VAE~\cite{wan2025wan}}
& \multicolumn{2}{c}{RepViTok (Ours)} \\
\cline{2-3}\cline{4-5}
& Easy & Hard
& Easy & Hard \\
\hline
Average$_{\mathrm{hor}=1}$
& 81.1 & 78.4
& \textbf{86.2} & \textbf{83.1} \\
Average$_{\mathrm{hor}=2}$
& 75.5 & 73.9
& \textbf{85.7} & \textbf{84.0} \\
Average$_{\mathrm{hor}=3}$
& 67.2 & 68.0
& \textbf{92.0} & \textbf{85.4} \\
\hline
Average$_{\mathrm{50\ Tasks}}$
& 78.0 & 76.0
& \textbf{86.6} & \textbf{83.1} \\
\hline
\end{tabular*}
\end{table*}

\noindent \textbf{World action modeling with latent actions.} We then study how latent actions should be incorporated into WAM pretraining in Table~\ref{tab:latent_actions}. The \textbf{w/o} baseline uses the RepViTok visual tokenizer but omits latent-action pretraining and trains the WAM directly on robot actions. \textbf{Joint Pred} trains the world model expert to predict video latents and action latents simultaneously through an additional prediction head, with the final action-latent slot padded by zero. \textbf{Two Stages} corresponds to our proposed design, in which the model is first pretrained on visual and latent action tokens and subsequently adapted to robot control.

\begin{table*}[!ht]
\centering
\caption{Ablation study on latent actions for world action modeling. We compare w/o latent actions, joint prediction, and two-stage latent action training.}
\label{tab:latent_actions}
\vspace{0.02in}
\small
\renewcommand{\arraystretch}{1.2}
\begin{tabular*}{\textwidth}{@{\extracolsep{\fill}}lcccc|cccc|c@{}}
\hline
\multirow{2}{*}{\textbf{Model}}
& \multicolumn{4}{c|}{Eval Seen} & \multicolumn{4}{c|}{Eval Unseen} & \multirow{2}{*}{PickFruit} \\
\cline{2-5}\cline{6-9}
& gFVD & PSNR & SSIM & OLS
& gFVD & PSNR & SSIM & OLS & \\
\hline
w/o
& 61.01 & 18.47 & 0.70 & 18.82
& 72.91 & 17.72 & 0.67 & 14.15 & 30\% \\
Joint Pred
& 94.25 & 15.24 & 0.61 & 18.52
& 98.77 & 15.09 & 0.55 & 15.22 & 20\% \\
\textbf{Two Stages (Ours)}
& \textbf{48.23} & \textbf{22.86} & \textbf{0.75} & \textbf{19.87}
& \textbf{58.83} & \textbf{19.93} & \textbf{0.74} & \textbf{16.98} & \textbf{50\%} \\
\hline
\end{tabular*}
\end{table*}

Joint Pred degrades visual dynamics, increasing gFVD to $94.25$ / $98.77$ and lowering both PSNR and SSIM, indicating that a single appended head entangles state prediction with action supervision. Two Stages achieves the best result on every metric, reducing gFVD to $48.23$ / $58.83$, improving PSNR to $22.86$ / $19.93$, and raising OLS to $19.87$ / $16.98$. The PickFruit result further confirms that the benefit is not limited to offline prediction: two-stage training improves closed-loop success to $50\%$, compared with $30\%$ without latent actions and $20\%$ for joint prediction. This gap suggests that pretraining on semantic latent actions provides a more transferable intermediate action representation, while direct robot-action supervision or joint prediction yields weaker executable behavior after adaptation. These results support decoupling latent-action pretraining from robot-action adaptation.

\begin{figure}[!ht]
\centering
\includegraphics[width=\linewidth]{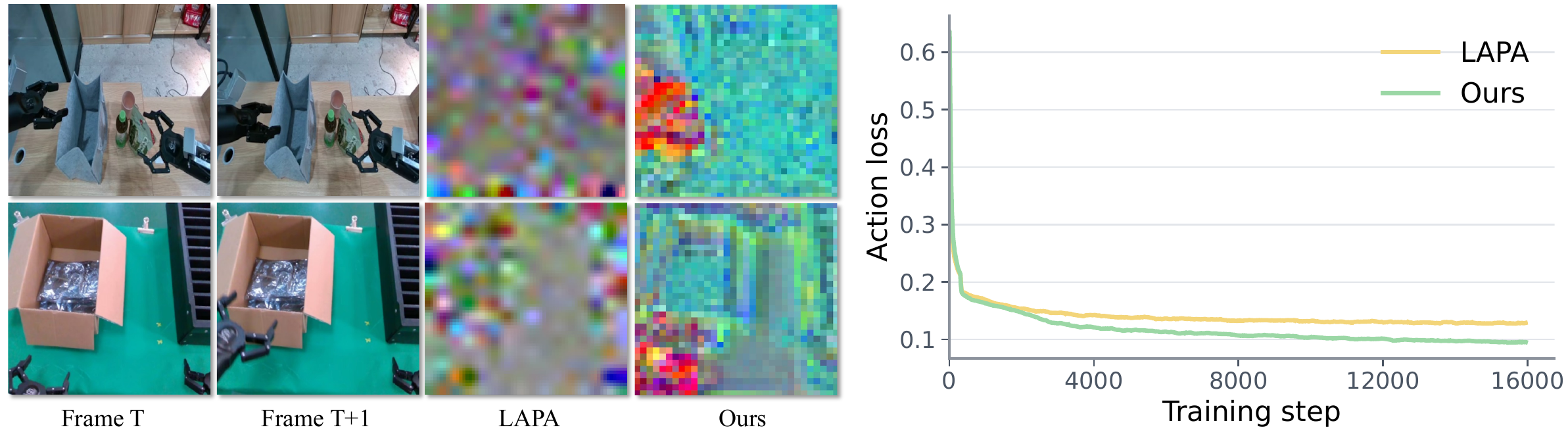}
\vspace{-0.2in}
\caption{Comparison of latent actions and their transfer to robot control. Left: action-latent visualizations from LAPA~\cite{ye2025latent} and RepViTok. Right: IDM loss with frozen action latents.}
\label{fig:latents}
\vspace{-0.2in}
\end{figure}

\noindent \textbf{Latent actions and transferring to robot control.} We further analyze whether the learned latent actions provide a useful bridge from visual dynamics to executable robot control in Figure~\ref{fig:latents}. The left side compares action-latent visualizations from LAPA~\cite{ye2025latent} and RepViTok. LAPA produces relatively diffuse responses, while our latent actions concentrate more clearly on manipulation-relevant changes, such as object displacement and contact-induced motion, indicating that RepViTok better captures the action-conditioned state transition. The right side evaluates transferability by freezing the learned action latents and training the same inverse dynamics model to decode robot actions from them. RepViTok achieves a lower IDM action loss than LAPA, showing that our latent actions are not only more visually aligned with motion but also easier to map into the robot-action space. Together, these results support the role of semantic latent actions as an effective intermediate representation for downstream control adaptation.

\begin{figure}[!ht]
\centering
\includegraphics[width=\linewidth]{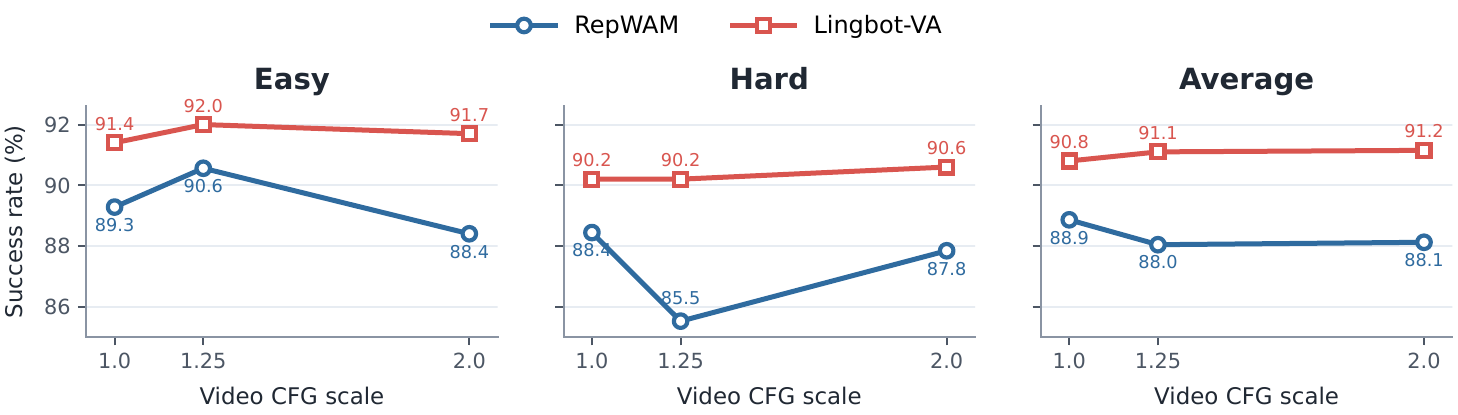}
\caption{Effect of video classifier-free guidance on RoboTwin 2.0 success rate. We compare our RepViTok-based WAM with Lingbot-VA~\cite{li2026causal} on Easy, Hard, and averaged success rates over video CFG scales $1.0$, $1.25$, and $2.0$.}
\label{fig:robotwin_cfg_curve}
\end{figure}

\paragraph{The effects of video classifier-free guidance.} In WAMs, action prediction generally does not require CFG, but the world model branch naturally inherits the dependence on CFG from video generation models. Given the stronger semantic structure of the RepViTok latent space, we examine whether this inherited dependence can be reduced by running inference with different video CFG scales. Figure~\ref{fig:robotwin_cfg_curve} compares our RepViTok-based WAM with Lingbot-VA~\cite{li2026causal} at scales $1.0$, $1.25$, and $2.0$. In our implementation, video CFG only affects the video denoising path, while action prediction remains unchanged.

For our model, the averaged success rate is highest at video CFG scale $1.0$, \ie, without additional CFG extrapolation. This indicates that RepViTok already provides a language-aligned visual-action space, making stronger video CFG unnecessary and sometimes harmful in harder randomized settings. It further shows that RepViTok weakens the inherited reliance on video CFG, allowing inference to use the conditional video branch alone when a CFG scale of $1.0$ is sufficient, which can reduce latency and activation memory in deployments that would otherwise evaluate an unconditional video branch.

\paragraph{Visual reconstruction results.} We evaluate the visual reconstruction results on ImageNet~\cite{deng2009imagenet} and UCF101~\cite{soomro2012ucf101} in Table~\ref{tab:tokenizer_reconstruction}. Across $256$ and $512$ resolutions, RepViTok remains competitive with WAN2.2 VAE in reconstruction quality, achieving higher ImageNet PSNR and SSIM and stronger UCF101 rFVD while delivering the downstream gains reported in Table~\ref{tab:tokenizer_ablation}.

\begin{table*}[!ht]
\centering
\caption{Reconstruction quality of different visual tokenizers on image and video benchmarks. We report rFID, PSNR, and SSIM on ImageNet at $256$ and $512$ resolutions, and rFVD, PSNR, and SSIM on UCF101 at $256$ and $512$ resolutions and $17$ frames.}
\label{tab:tokenizer_reconstruction}
\vspace{0.05in}
\small
\renewcommand{\arraystretch}{1.45}
\setlength{\tabcolsep}{3pt}
\begin{tabular*}{\textwidth}{@{\extracolsep{\fill}}lcccccccccccc@{}}
\hline
& \multicolumn{3}{c}{ImageNet 256}
& \multicolumn{3}{c}{ImageNet 512}
& \multicolumn{3}{c}{UCF101 256$\times$17}
& \multicolumn{3}{c}{UCF101 512$\times$17} \\
\cline{2-4}\cline{5-7}\cline{8-10}\cline{11-13}
\textbf{Model}
& rFID & PSNR & SSIM
& rFID & PSNR & SSIM
& rFVD & PSNR & SSIM
& rFVD & PSNR & SSIM \\
\hline
WAN2.2 VAE~\cite{wan2025wan}
& \textbf{0.50} & 28.16 & 0.87
& \textbf{0.20} & 30.48 & 0.90
& 4.28 & \textbf{36.61} & \textbf{0.98}
& 0.68 & \textbf{41.45} & \textbf{0.99} \\
ViTok
& 0.96 & 28.65 & \textbf{0.89}
& 0.24 & 30.77 & \textbf{0.92}
& 1.23 & 35.52 & 0.97
& \textbf{0.16} & 40.68 & 0.98 \\
\textbf{RepViTok}
& 0.80 & \textbf{28.90} & \textbf{0.89}
& 0.23 & \textbf{31.00} & \textbf{0.92}
& \textbf{1.09} & 36.03 & 0.97
& \textbf{0.16} & 41.12 & 0.98 \\
\hline
\end{tabular*}
\end{table*}

\begin{figure*}[!ht]
\centering
\includegraphics[width=0.98\textwidth]{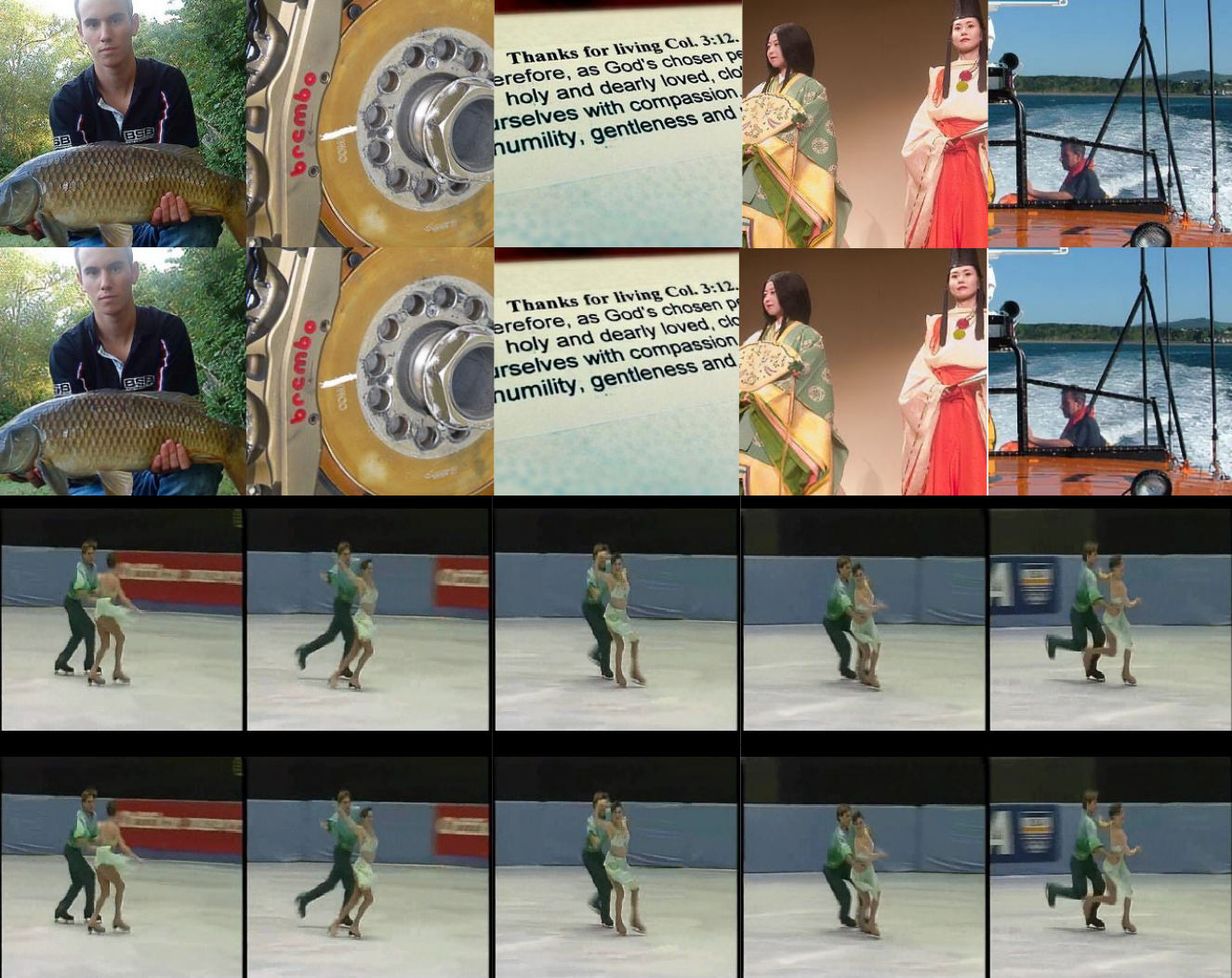}
\caption{Qualitative reconstruction examples on ImageNet and UCF101. RepViTok preserves semantic details in images and maintains temporal consistency in videos.}
\label{fig:reconstruction}
\vspace{-0.15in}
\end{figure*}

The qualitative reconstructions in Figure~\ref{fig:reconstruction} complement the quantitative results by showing what RepViTok preserves in individual examples from ImageNet and UCF101. On ImageNet images, RepViTok maintains both low-level appearance and semantic identity: faces remain recognizable, object boundaries stay sharp, and text regions preserve coherent strokes and layouts rather than collapsing into blurry textures. On UCF101 videos, RepViTok further preserves temporal consistency across frames, keeping moving actors and objects stable even in clips with large body motion and fast foreground changes. These examples indicate that the semantic alignment objective preserves competitive reconstruction quality despite tradeoffs on some fidelity metrics. More importantly, it helps the tokenizer retain visually and semantically important details in both static images and dynamic videos.

\section{Conclusion}
\label{sec:conclusion}

We presented \system, a representation-centric world action model built on semantic visual-action tokenization. Rather than relying on reconstruction-oriented video latents inherited from pretrained generation backbones, \system aligns visual latents with a frozen visual foundation model and learns latent actions as transitions between visual states. With this semantic representation, we train a causal diffusion transformer to jointly model instruction-conditioned future observations and the latent actions that realize them, and then adapt these dynamics to embodiment-specific robot control with demonstrations. Experiments on Franka dual-arm real-world manipulation and RoboTwin 2.0, together with tokenizer and latent-action ablations, show that semantic visual-action alignment improves visual dynamics, action prediction, and closed-loop execution. These findings highlight the significance of representation design for WAMs, pointing toward more interpretable, transferable, and effective robot manipulation.

For future work, we will scale WAM pretraining beyond the robotics-domain videos by leveraging large-scale internet videos, especially egocentric human videos. This could broaden the range of behaviors and interaction patterns available during pretraining.

\bibliographystyle{plainnat}
\bibliography{main}

\end{document}